\documentclass[11pt,letterpaper]{llncs}
\usepackage{amsmath}
\usepackage{amssymb}
\usepackage{graphicx}
\usepackage{subcaption}
\usepackage{marvosym}
\usepackage{url}
\usepackage{fancyhdr}
\usepackage{booktabs}
\usepackage{tikz}
\usepackage[numbers,sort&compress]{natbib}

\usepackage{geometry}
\newcommand{\keywords}[1]{\par\addvspace\baselineskip
\noindent\keywordname\enspace\ignorespaces#1}

\begin{document}

%\mainmatter  % start of an individual contribution

\title{\LARGE{AR Overlay: Training Image Pose Estimation on Curved Surface in a Synthetic Way}}

%\titlerunning{Lecture Notes in Computer Science: Authors' Instructions}

\author{\large{Sining Huang\inst{1}, Yukun Song\inst{1}, Yixiao Kang\inst{1}, Chang Yu\inst{2}}}
\institute{\large{University of California - Berkeley at Berkeley, CA 94720, USA \inst{1} \\ \large{Northeastern University at Boston, MA 02115, USA \inst{2}}}}

%\author{Alfred Hofmann%
%\thanks{Please note that the LNCS Editorial assumes that all authors have used
%the western naming convention, with given names preceding surnames. This determines
%the structure of the names in the running heads and the author index.}%
%\and Ursula Barth\and Ingrid Haas\and Frank Holzwarth\and\\
%Anna Kramer\and Leonie Kunz\and Christine Rei\ss\and\\
%Nicole Sator\and Erika Siebert-Cole\and Peter Stra\ss er}

%\institute{Springer-Verlag, Computer Science Editorial,\\
%Tiergartenstr. 17, 69121 Heidelberg, Germany\\
%\url{http://www.springer.com/lncs}}


%\toctitle{Lecture Notes in Computer Science}
%\tocauthor{Authors' Instructions}

\maketitle

\begin{abstract}
\textit{In the field of spatial computing, one of the most essential tasks is the pose estimation of 3D objects. While rigid transformations of arbitrary 3D objects are relatively hard to detect due to varying environments introducing factors like insufficient lighting or even occlusion, objects with pre-defined shapes are often easy to track, leveraging geometric constraints. Curved images, with flexible dimensions but a confined shape, are essential shapes often targeted in 3D tracking. 
Traditionally, proprietary algorithms often require specific curvature measures as the input along with the original flattened images to enable pose estimation for a single image target. In this paper, we propose a pipeline that can detect several logo images simultaneously and only requires the original images as the input, unlocking more effects in downstream fields such as Augmented Reality (AR).}
\end{abstract}

%\begin{abstract}
%The abstract should summarize the contents of the paper and should
%contain at least 70 and at most 150 words. It should be written using the
%\emph{abstract} environment.
%\keywords{We would like to encourage you to list your keywords within
%the abstract section}
%\end{abstract}

\section{Introduction}
Augmented reality, with a primary focus on weaving intellectual interfaces and digital arts \cite{kang2022tie} into people's real lives in a seamless way, has become one important research area as it involves multiple modalities in computation and elicits improvement needs in various fields \cite{yang2022retargeting,zhu2023demonstration}.
Our objective is to develop an algorithm for tracking curved images by accurately estimating the diameter, transition, and rotation of cylinder-like objects in real time. This algorithm has significant potential for applications such as augmented reality in retail, where it overlays digital information onto curved products, as well as in robotic grasping and human-robot interaction for learning demonstrations. The process involves using YOLOv8 \cite{yolo_2016,Jocher_Ultralytics_YOLO_2023,Ma_2024} for logo detection, Convolutional Neural Network (CNN) ~\cite{xiang2024neural} for diameter estimation, and SIFT \cite{sift} for feature extraction, culminating in pose estimation through solving a Perspective-n-Point (PnP) problem ~\cite{zhang_flexible_2000}. This approach enhances the algorithm's capability to handle complex real-world scenarios effectively.

\begin{figure}[ht]
  \centering
  \includegraphics[width=0.2\linewidth]{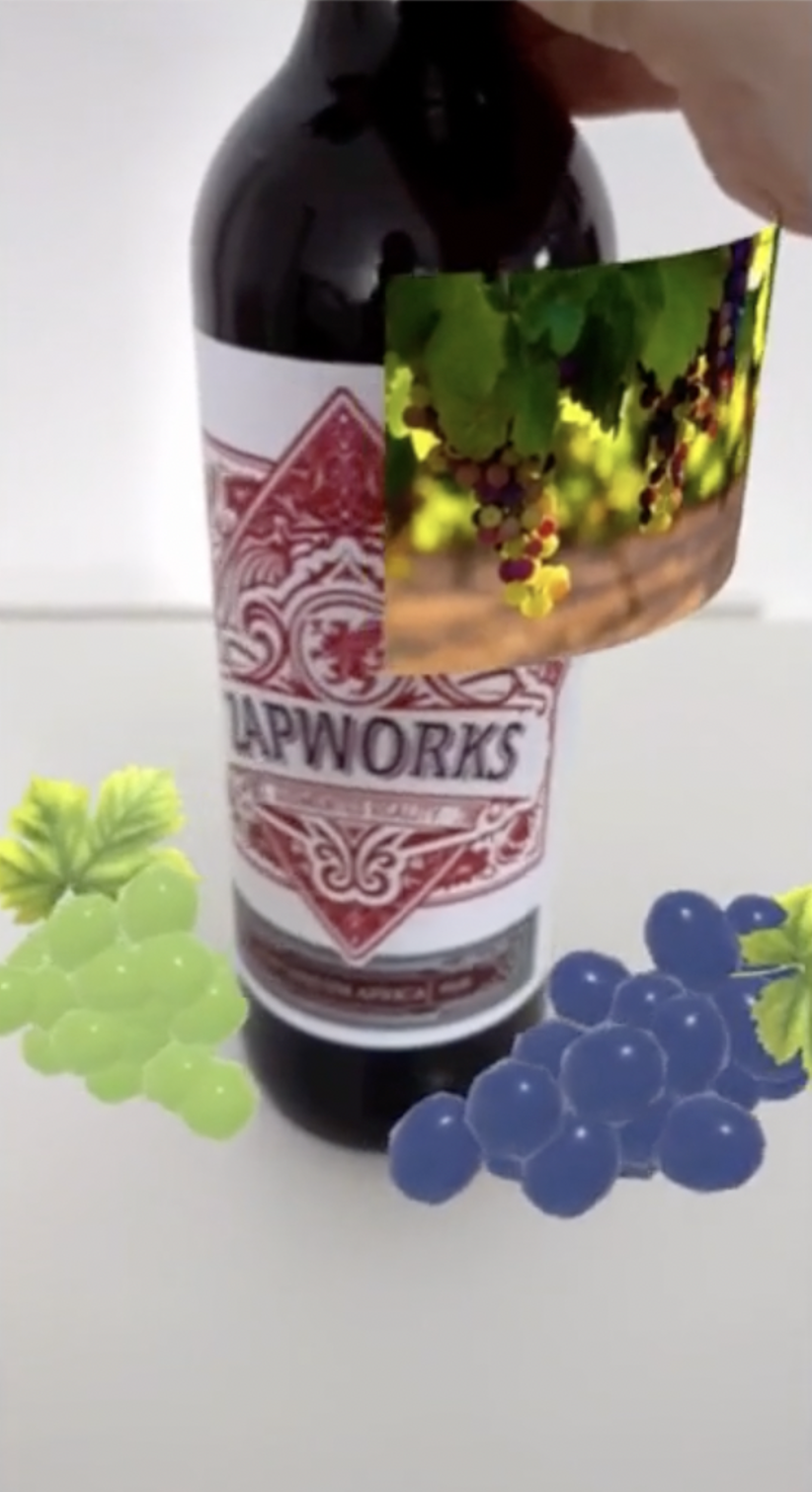}
  \caption{AR Effects shown by ZapWorks’ single-image pose-estimation algorithm}
  \label{fig:zapworks}
\end{figure}
ZapWorks currently provides a basic, proprietary tool that supports curved image tracking ~\cite{zapworks_curved}, as depicted in Figure ~\ref{fig:zapworks}. Given its closed-source nature, our main objective is to develop a comparable model using convolutional networks and traditional estimation methods ~\cite{malik1997computing,dan_multiple_2024}. A significant limitation of ZapWorks' solution is its capacity to detect only a single, predetermined image, such as the logo on a wine bottle shown in Figure ~\ref{fig:zapworks}. To address this, we want to enhance our model's ability to recognize multiple pre-trained logos autonomously, without the need to input specific logo IDs for each usage. Our ultimate aim is to evolve this model into a few-shot, closed-set algorithm, similar to those found in other pose-estimation frameworks ~\cite{feng2023dttd,he2022fs6d}, thus vastly improving its utility and efficacy in a variety of real-world scenarios.

Our work provides the following contributions: \begin{enumerate}
    \item We synthesize the training \& evaluation dataset using Blender, comprising over 20,000 images, enabling model training for curved image detection, tracking, and diameter estimation.
  \item We introduce a novel framework for 6D pose estimation of curved images, leveraging a combination of SIFT and solving a PnP problem.
  \item We propose a novel CNN-based architecture tailored for estimating the curvature of images (i.e., the diameter of the underlying cylinder) based on the bounding box of the logo image.
\end{enumerate}

\section{Related Works}

\subsection{General Object Detection}
In the field of general object detection, methodologies are categorized into one-stage and two-stage detectors. Two-stage detectors, such as R-CNN ~\cite{girshick_rich_2014} and its variants Fast R-CNN ~\cite{girshick_fast_2015} and Faster R-CNN ~\cite{ren_faster_2016}, initially generate region proposals that likely contain objects, which are subsequently classified and refined in terms of bounding box coordinates ~\cite{tao2023sqba,tao2023mlad}. These models are noted for their high accuracy, especially in complex scenes with small or overlapping objects ~\cite{xin2024vmt,xin2024mmap}, but tend to operate slower due to their two-phased approach. On the other hand, one-stage detectors like You Only Look Once (YOLO) \cite{yolo_2016} and Single Shot Multibox Detector (SSD) ~\cite{liu_ssd_2016} simplify the detection process by directly predicting bounding boxes and class probabilities in a single step, trading some accuracy for significant gains in speed and efficiency. For instance, typical applications such as a self-guided retail checkout process developed by Tan et al. \cite{tan2024enhanced} showcases the real-time advantage of one-stage detectors like YOLO. Furthermore, an exemplar work by Dang et al. \cite{dang2024realtime} demonstrates the strong adaptability of the YOLO. Dang et al.'s work not only fine-tuned the bounding box output using new datasets, but also presented the high-precision classification of small items like pills using the YOLO network. Furthermore, their research, which focuses on recognition targets that share same shape and color (typical barriers to manual classifications), proves the strong generalization ability of the YOLO network and sheds light on how it can be utilized in more complex real-life scenarios.

\subsection{6D Object Pose Estimation}
In the field of pose estimation, methodologies are categorized into traditional geometric techniques and advanced deep-learning approaches ~\cite{dan_evaluation_2024,article,WANG2022106551,202408.0927}. Traditional approaches often involve feature extraction and matching with 3D models using Perspective-n-Point (PnP) solutions or Iterative Closest Point (ICP) ~\cite{besl_method_1992,xiang2023hybrid} algorithms. Modern deep learning techniques, such as PoseCNN ~\cite{xiang_posecnn_2018,Li2024.09.08.24313212}, leverage convolutional neural networks to predict object pose directly from RGB images, enhancing efficiency and scalability. Hybrid methods like DenseFusion ~\cite{wang2019densefusion} and PVNet ~\cite{peng_pvnet_2018} integrate RGB data with depth information, improving accuracy and robustness in complex scenes ~\cite{fan2024towards,liu2024enhancing,liu2024deeplung}. These methods can effectively handle occlusions and varying lighting conditions, crucial for applications in augmented reality, robotics, and autonomous navigation.

\subsection{Convolutional Neural Networks}
CNNs have been adapted for various complex image-processing tasks by modifying the embedding layers. For instance, in style transfer, CNNs are engineered to encode style and content features separately, combining them to produce new, artistic images. In object detection, adjustments to the embedding layer integrate with region proposal networks to enhance detection accuracy ~\cite{tang2024research}. Such modifications improve the network's ability to extract detailed features crucial for tasks like facial recognition or medical imaging anomaly detection ~\cite{he_deep_2015,song_deep_2019}. These adaptations underscore the flexibility of CNNs to meet specific operational demands across diverse applications ~\cite{li2024ddn}.

\section{Dataset}
Given the specificity of our research needs, we opted to create a synthetic dataset using Blender, as no suitable dataset was available online. This dataset includes 20,000 images, generated from 20 different target images, each affixed to cylinders of various diameters and heights. For each cylinder, over 1000 image frames were produced to capture diverse perspectives. Note that, in this dataset, the term 'target image' specifically refers to any image used to represent a logo.
\begin{figure}[ht]
  \centering
  \includegraphics[width=1\linewidth]{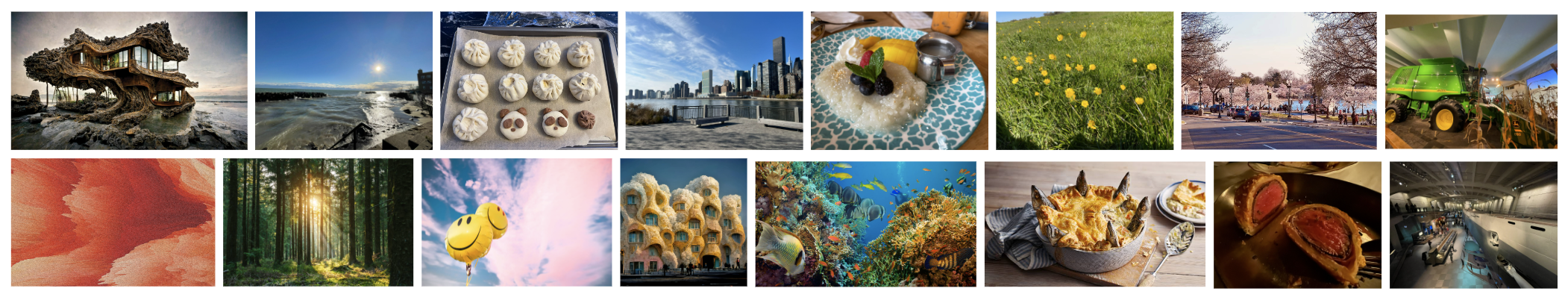}
  \caption{Collection of 20 different target images (some are removed due to anonymization)}
  \label{fig:logo_collect}
\end{figure}

\begin{figure}[ht]
  \centering
  \includegraphics[width=1\linewidth]{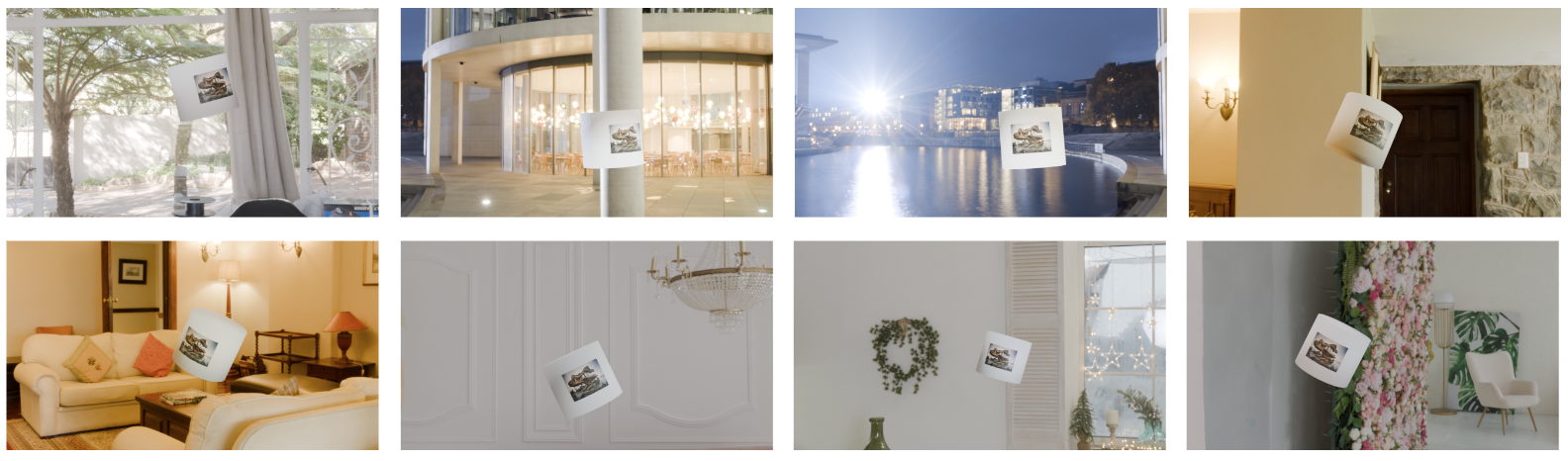}
  \caption{Sample of Cylinder Object Across Varied Backgrounds}
  \label{fig:background_collect}
\end{figure}

\subsection{Dataset Generation}
For each iteration, we selected one of the 20 target images in sequence to ensure a comprehensive evaluation across all variations (see all target images in Fig. ~\ref{fig:logo_collect}). To enhance the model’s adaptability to different backgrounds, we incorporated 15 distinct 360-degree background images. For example, the cylinder with the Cal logo can be placed into various backgrounds as shown in Fig. ~\ref{fig:background_collect}. 

Additionally, we generated cylinders with diameters ranging from one to two times the width of the attached image, placing the target image on the side of the cylinder. The camera was maneuvered around the cylinder to capture images from multiple angles, ensuring that the target image remained visible regardless of perspective. To simulate realistic conditions, such as objects captured off-center as might occur in real-world usage, we varied the cylinder’s position within each image. This methodological diversity is crucial for training our model to accurately detect targets under a variety of conditions.

During the synthesis process, we recorded additional training data, including the label width, label height, and the camera’s intrinsic matrix, which are essential for applying the Scale Invariant Feature Transform (SIFT) and calculating the transformation matrix. We documented the ground truth data for each synthesized image, detailing the cylinder’s relative position and rotation, measured by Euler angles, in relation to the camera, as well as its diameter. Additionally, we recorded the camera's intrinsic matrix, which is a fundamental component in camera calibration. The intrinsic matrix is a 3x3 matrix that transforms 3D coordinates from the camera's view into 2D image coordinates, facilitating the conversion of real-world measurements to pixel measurements ~\cite{tan2024editable}. It includes parameters such as the focal lengths ($f_x$ and $f_y$), which determine the scaling in the x and y directions, the skew coefficient ($s$), which accounts for any non-rectangularity between the x and y pixel axes, and the principal point coordinates ($c_x$ and $c_y$), which indicate the intersection of the optical axis with the image plane. This matrix accurately projects 3D points onto the 2D image plane, allowing for precise image formation and analysis ~\cite{zhou2024evaluating}.

\begin{figure}[ht]
    \centering
    \begin{subfigure}{0.44\textwidth}
        \centering
        \includegraphics[width=\linewidth]{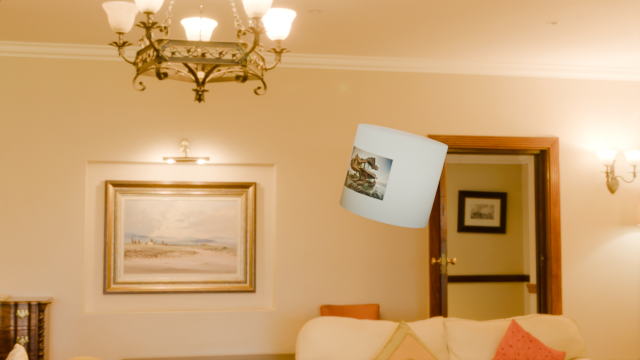}
        \caption{Input image}
        \label{fig:input-image}
    \end{subfigure}
    \hfill
    \begin{subfigure}{0.51\textwidth}
        \centering
        \scriptsize
        \begin{tabular}{ll}
            \toprule
            \textbf{Relative Position} & $\langle 0.1704, -1.1875, 0.203 \rangle$ \\ 
            \textbf{Relative Rotation (Euler)} & $\langle -1.6166, -0.1995, -0.1264 \rangle$ \\ 
            \midrule
            \textbf{Cylinder Diameter} & 1.64 \\ 
            \textbf{Label Width} & 1.3333 \\ 
            \textbf{Label Height} & 1.0 \\ 
            \midrule
            \textbf{Camera Intrinsic Matrix} & 
            $\begin{bmatrix}
            2670 & 0 & 960 \\
            1 & 2250 & 540 \\
            1 & 0 & 1
            \end{bmatrix}$ \\ 
            \bottomrule
        \end{tabular}
        \caption{Corresponding data}
        \label{fig:corresponding-data}
    \end{subfigure}
    \caption{Synthesized dataset consisting of paired images and corresponding data examples}
    \label{fig:datapair}
\end{figure}

For clarity and consistency, we define the curvature of the image target as the ratio between the cylinder’s diameter and the height of the image target (HoI). Fig. ~\ref{fig:datapair} illustrates a sample image pair from the dataset, accompanied by its relevant training and testing data, showcasing the dataset's capacity to thoroughly evaluate our model’s performance.

As the dataset randomly generated backgrounds and cylinder's diameter, we split the first 90\% of the data into training datasets while the remaining 10\% as the validation dataset. The split in the order can avoid backtracking the image index. 

\section{Method}
\begin{figure*}[ht]
  \centering
  \includegraphics[width=\linewidth]{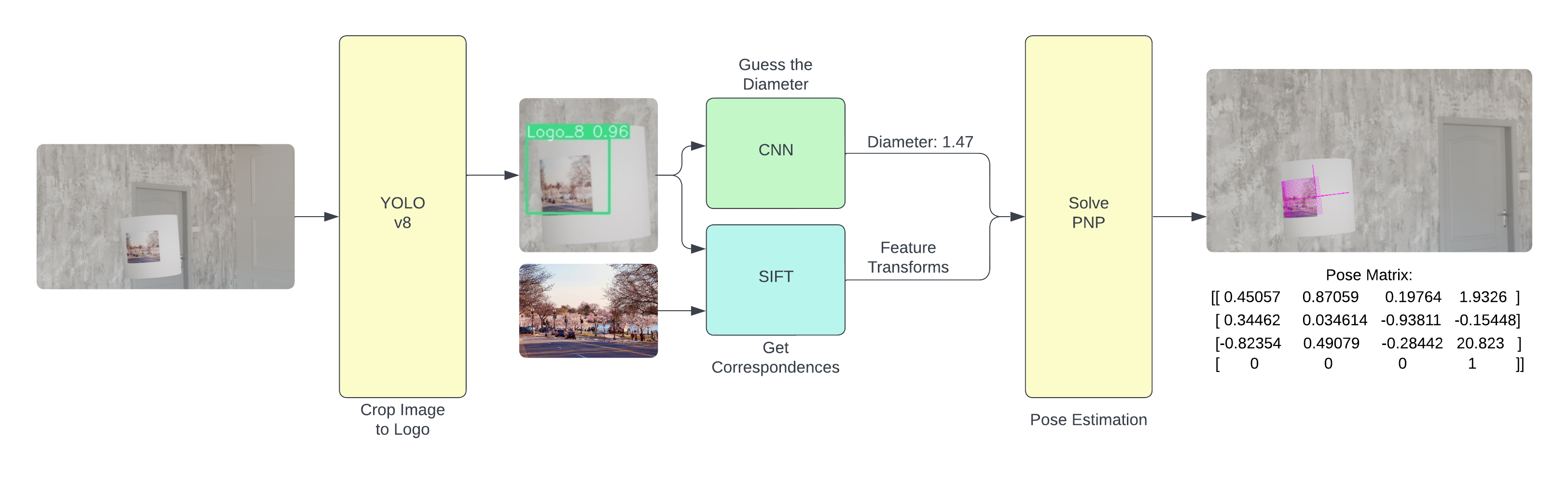}
  \caption{Our pipeline of curved-image pose-estimation. The pipeline contains a fine-tuned YOLOv8 network \cite{yolo_2016}, a self-trained CNN, a feature matching algorithm using SIFT ~\cite{sift}, and an algorithm for PnP ~\cite{zhang_flexible_2000} pose computation}
  \label{fig:architecture}
\end{figure*}

Generally, we divide the problem into four steps, including 1) detecting the bounding box of the image, 2) estimating the curvature of the image, 3) finding correspondences between the flattened image target and the input image capture, and 4) estimating the pose of the image. Fig. ~\ref{fig:architecture} presents the pipeline in detail using a training image as an example. The pose of the image is hence referring to the pose of the cylinder, where the image is always attached to (and centered to) the negative y-direction of the cylinder, with the top of the cylinder being the positive z-direction and the right being the positive x-direction.

\subsection{Image Classification \& Detection}
\begin{figure*}[h]
  \centering
  \includegraphics[width=0.75\linewidth]{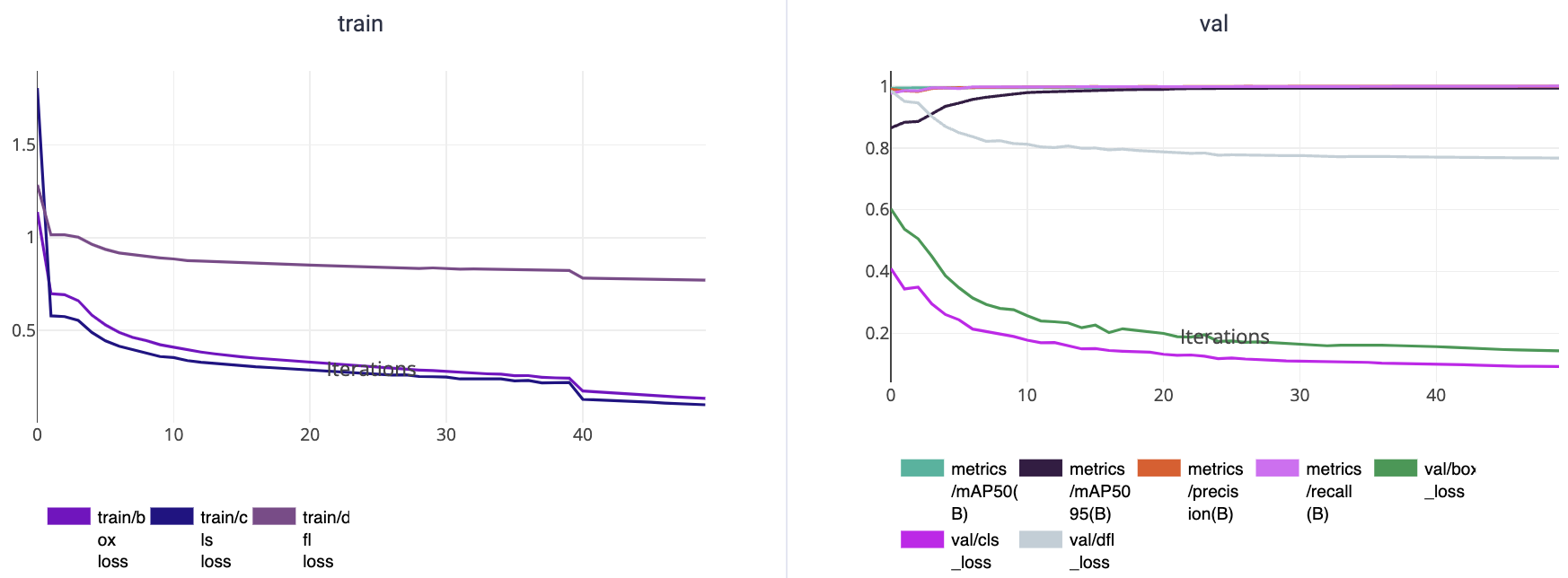}
  \caption{The training loss (left) and the validation loss (right) for fine-tuning the YOLOv8 network.}
  \label{fig:yolo_loss}
\end{figure*}
For step 1), inspired by Tan et al. \cite{tan2024enhanced}, who leverages the detection head structure from YOLOv8 to conduct high-precision object recognition, we fine-tune a pre-trained YOLOv8 network using the small checkpoint (yolov8s.pt) provided by prior literature to detect the bounding boxes for the images. With a 50-epoch training process using 921 training captures per image target (20 image targets in total in our dataset), the validation box\_loss, referring to the regression Complete Intersection over Union (CIoU) loss \cite{zheng_enhancing_2021}, dropped to 0.14355 as shown in Fig. ~\ref{fig:yolo_loss}, representing a decent prediction result. Notice that the YOLOv8 network also performs the image classification in this step.
\subsection{Curvature Estimation}
For step 2), we propose a new convolution-based network for predicting the curvature of the image. We experimented with two differently-sized networks with both the Huber loss function ($\delta = 0.4$) ~\cite{Shen2024Harnessing,zhang2024cunetunetarchitectureefficient} and the traditional mean squared error (MSE) loss function ~\cite{zheng2024identification}. Both networks include 3 convolution layers, each followed by a max-pooling layer with kernel size $2*2$, and two fully connected layers, where one layer encodes the kernel output from the prior convolutional layer into vector embedding and the second layer fully connects the first layers' output and combines into one single output value as the curvature prediction. All layers use the Rectified Linear Unit (ReLU) activation function, which is a standard usage \cite{agarap_deep_2019} for training convolutional networks. Specifically, the small-sized network has convolutional layers with kernels numbers 32, 64, 64, with the kernel size being $5*5, 3*3, 4*4$, followed by a 64-dimension vector embedding layer, while the large-sized network has convolutional kernel numbers being 32, 64, 128, 128, with the kernel size being $5*5, 3*3, 3*3$, and followed by a 128-dimension vector embedding layer. In total, the small-sized network contains 189,057 parameters while the large-sized network contains 684,865 parameters. 

All training processes were set to run for 50 epochs, and an early stopping mechanism was applied when the validation loss stopped decreasing for consecutively 4 epochs. With training results shown in Table ~\ref{tab:model_performance}, the small-sized model with the Huber loss got the best performance. In the following section of evaluation and result, we evaluate both the Small-Size-Huber-Loss model and the Small-Size-MSE-Loss model to give a better indication of the effects of the two different loss functions.

\begin{table}[ht]
\centering
\caption{Value of metrics during training}
\label{tab:model_performance}
\begin{tabular}{l@{\hspace{10pt}}c@{\hspace{10pt}}c@{\hspace{10pt}}c}
\toprule
\textbf{Model Type} & \textbf{Best Epoch} & \textbf{Val Loss Huber ($\delta = 0.4$)} & \textbf{Val Loss MSE} \\
\midrule
\textbf{Small-Size-Huber-Loss} & 24 & \textbf{0.0266} & \textbf{0.0787} \\
\textbf{Large-Size-Huber-Loss} & 5 & 0.0523 & 0.1175 \\
\textbf{Small-Size-MSE-Loss} & 8 & 0.0466 & 0.1042 \\
\textbf{Large-Size-MSE-Loss} & 4 & 0.0640 & 0.1522 
\end{tabular}
\end{table}

\subsection{Finding Feature Correspondences}
In this step, we utilize the SIFT algorithm to find the image correspondences between the flat image target and the image capture. The Fast Library for Approximate Nearest Neighbors (FLANN) \cite{muja_flann_2009} was then utilized to find the matches between the feature descriptors using the K-nearest-neighbors method \cite{kaplan_nonparametric_1958}. A ratio of 0.95 was then used in the ratio test to filter the bad matches. 

\subsection{Pose Estimation}
Given the curvature prediction, we can easily map the original points in the flat image target onto the curved surface in the cylinders' coordinate system using the following formula, where $r$ stands for the radius of the cylinder (notice that all image coordinates are using the unit of HoI.):
\begin{align*}
    \theta &= X_{\text{Target}} - (\frac{TargetWidth}{2}) \\
    x_{\text{Cyl}} &= r*sin(\theta) \\
    y_{\text{Cyl}} &= - r*cos(\theta) \\
    z_{\text{Cyl}} &= (\frac{TargetHeight}{2}) - Y_{\text{Target}} \\
\end{align*}

Then, given the correspondences between the extracted image features, we can perform the PnP pose computation as our last step in the pipeline. To make our pipeline more robust against the errors generated during the SIFT feature extraction process (caused by the non-linear image distortion on the curved surface), we utilize the RANSAC method \cite{derpanis_overview_nodate} to solve the PnP problem and get the pose estimation.

% Time
\section{Evaluation \& Result}

\begin{figure}[h]
  \centering
\includegraphics[width=0.32\linewidth]{imgs/example_image_input.png}
\includegraphics[width=0.32\linewidth]{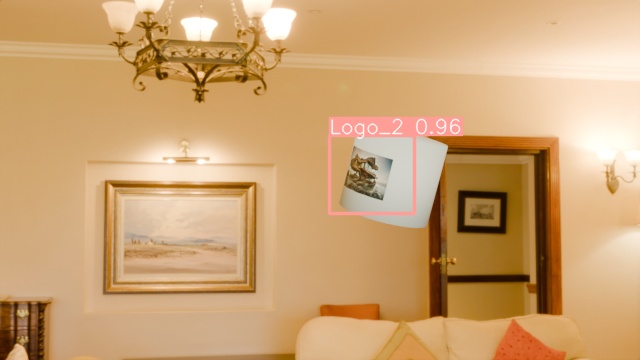}
\includegraphics[width=0.32\linewidth]{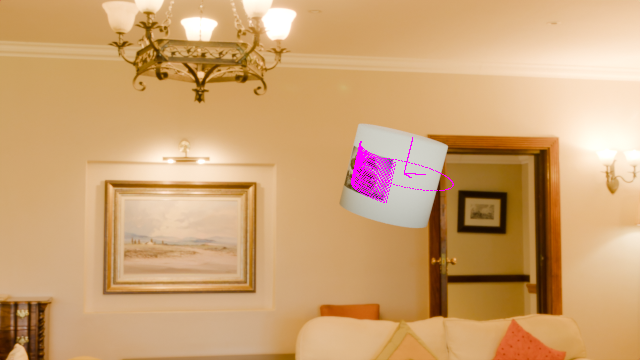}
\caption{Original Image (left), Yolo Detection (middle), and pose estimation (right) result of our pipeline.}
\label{fig:prediction}
\end{figure}

Our pipeline takes only the target image and the context image as input and accurately predicts the 6D pose (rotation and translation) of the target image using a hypothesized cylindric shape with a predicted diameter. Fig. ~\ref{fig:prediction} demonstrates the visual result of our prediction. Notice that in real use cases, the cylindric shape that the target image curls around does not necessarily need to appear in the context image. Here in the figure, the white cylinder is generated for illustration and comparison purposes only. In addition, we also conducted a quantitative analysis of our models as shown in Table ~\ref{tab:detailed_metrics}. Using the Small-Size-Huber-Loss and the Small-Size-MSE-Loss model, we can see the significant difference in the error of diameter estimation. The rotation error (based on the inner product of unit quaternions, $\Phi_6 \equiv 2 * \Phi_3$ in ~\cite{huynh_metrics_2009,zhang2024ratt,10.1145/3331184.3331244}), as well as the translation error, are also provided as a reference. Similar methods involving efficient prediction models for different tasks can be found in works such as ~\cite{zhang2024prototypical,ma2024transformer,fan2024advanced,xiang2023landing}, where data-efficient reinforcement learning, adaptive feature generation ~\cite{zhang2024dynamic,zhang2024tifg}, and reasoning structures are utilized for optimal results.

\begin{table}[h]
\centering
\caption{Quantitative analysis of the pipeline performance with different models for curvature estimation}
\begin{tabular}{l c c}
\toprule
\multicolumn{1}{c}{}& \multicolumn{1}{c}{\textbf{Small-Size}} & \multicolumn{1}{c}{\textbf{Small-Size}} \\
 \multicolumn{1}{c}{\textbf{Metrics}} & \multicolumn{1}{c}{\textbf{MSE-Loss Model}} & \multicolumn{1}{c}{\textbf{Huber-Loss Model}} \\ 
\midrule
\textbf{YOLO Success Rate} & 0.967 & 0.967 \\
\textbf{IoU} & 0.963 ± 0.060 & 0.963 ± 0.060 \\
\midrule
\textbf{Time Taken (s)} & 0.635 ± 0.156 & 0.626 ± 0.152 \\
\textbf{Diameter Error (HoI)} & 0.247 ± 0.208 & 0.176 ± 0.157 \\
\textbf{Rotation Error (rad)} & 2.524 ± 0.557 & 2.521 ± 0.558 \\
\textbf{Translation Error (HoI)} & 50.445 ± 19.109 & 50.731 ± 19.031 \\
\bottomrule
\end{tabular}
\label{tab:detailed_metrics}
\end{table}
\section{Discussion, Limitation, and Future Works}
Our pipeline provides the flexibility to simultaneously be compatible with multiple target images, solving the scalability issue in existing public methods \cite{zapworks_curved}. Specifically, considering the dataset generation process, our pipeline has the ability to complement the insufficient 3D clues that a single planar image may exhibit, which improves its compatibility in terms of target image categories. Similar processes can be applied to and enhance the performance of existing applications \cite{kang2021_simultaneous,curved_objects_tracking,deng2024incremental,weng2024leveraging}, which is not limited to the field of AR. For instance, eye-tracking algorithms \cite{eye_tracking,Weng202404,Weng2024} can also take similar approaches in addition to the machine-learning-primary architecture to refine their results, which could be applied to the manufacturing of head-mounted displays \cite{song_going_2023,song_looking_2024}. Applications in other fields, such as tracking and classifying irregular-shaped objects in real-time checkout systems mentioned in Tan et al.'s work \cite{tan2024enhanced} or general-purpose region tracking \cite{bascle_region_1995,deng2024multi}, can also benefit from implementing an additional and synchronous layer of pose estimation.

During training, we found that Huber loss outperformed Mean Squared Error (MSE) loss in our curvature estimation model. This points to potential improvements by exploring alternative loss functions and refining the model's architecture for optimal performance. Future efforts can focus on enhancing the model's real-time performance, integrating advanced machine learning techniques like reinforcement learning or attention mechanisms ~\cite{xubo_2024_12684615,jin2024online}, and expanding the dataset to include a wider variety of shapes and textures. Additionally, we aim to further refine the curvature estimation model, optimizing its architecture for better accuracy and efficiency, similar to the optimization strategies discussed in \cite{zhao2024optimization,Weng202406,qiao2024robust} for self-supervised learning.

Additionally, though our approach to curved image pose estimation shows promise, it has specific limitations that suggest avenues for future research. One key limitation is related to our dataset. Minor errors introduced by Blender's rendering optimization can cause the intrinsic matrix to inaccurately reflect the camera mapping, leading to discrepancies in pose estimation. A possible way that future work could address this issue is to generate datasets with a white-box rendering pipeline for improved accuracy.

By addressing these limitations and exploring these future directions, we hope to advance the field of curved image pose estimation, with significant implications for augmented reality and industrial applications.

\section{Conclusion}
In this paper, we have presented a comprehensive solution to the challenging problem of image pose estimation on a curved surface, offering both a concrete and scalable methodology and an in-depth analysis. Our work is grounded in the creation of a large synthetic dataset consisting of over 20,000 images, which serves as a foundation for training and evaluating our proposed models. We introduced a robust pipeline for 6D pose estimation of curved images, effectively leveraging a combination of YOLOv8 for object detection and a novel Convolutional Neural Network (CNN) architecture tailored for curvature estimation.

Our approach addresses the limitations of traditional methods that are restricted to single-image detection. By enabling the simultaneous detection of multiple target images, our pipeline significantly enhances the capability of pose estimation systems, paving the way for new possibilities in a variety of downstream applications. This is particularly impactful in the field of AR, where our model's ability to accurately overlay digital content on physically curved surfaces enhances user engagement and interaction. Additionally, our framework holds promise for broader industrial applications, including robotic manipulation and human-robot interaction, where precise tracking and pose estimation are critical. 

\renewcommand{\bibname}{References}
\bibliographystyle{IEEEtranN}
\bibliography{ref}

\end{document}